\documentclass{article} 
\usepackage[preprint]{colm2026_conference}

\usepackage{microtype}
\usepackage{hyperref}
\usepackage{url}
\usepackage{booktabs}
\usepackage{lineno}
\usepackage{float} 

\usepackage{latexsym}
\usepackage{graphicx}
\usepackage{amsmath}
\usepackage{amssymb}
\usepackage{mathtools}
\usepackage{amsthm}
\usepackage[edges]{forest}
\usepackage{tikz}
\usepackage{xurl}
\usepackage[capitalize,noabbrev]{cleveref}
\usepackage{xcolor}

\usepackage{wrapfig}
\usepackage{enumitem}
\usepackage{tabularx}
\usepackage{booktabs}

\usepackage[T1]{fontenc}
\usepackage[utf8]{inputenc}

\definecolor{darkblue}{rgb}{0, 0, 0.5}
\hypersetup{
  colorlinks=true,
  citecolor=darkblue,
  linkcolor=darkblue,
  urlcolor=darkblue
}

\newcommand*\BC[1]{%
\begin{tikzpicture}[baseline=(C.base)]
\node[draw,circle,fill=black,inner sep=0.2pt](C) {\textcolor{white}{#1}};
\end{tikzpicture}}

\theoremstyle{plain}

\theoremstyle{definition}

\theoremstyle{remark}

\definecolor{paired-light-blue}{RGB}{198, 219, 239}
\definecolor{paired-dark-blue}{RGB}{49, 130, 188}
\definecolor{paired-light-orange}{RGB}{251, 208, 162}
\definecolor{paired-dark-orange}{RGB}{230, 85, 12}
\definecolor{paired-light-green}{RGB}{199, 233, 193}
\definecolor{paired-dark-green}{RGB}{49, 163, 83}
\definecolor{paired-light-purple}{RGB}{218, 218, 235}
\definecolor{paired-dark-purple}{RGB}{117, 107, 176}
\definecolor{paired-light-gray}{RGB}{217, 217, 217}
\definecolor{paired-dark-gray}{RGB}{99, 99, 99}
\definecolor{paired-light-pink}{RGB}{222, 158, 214}
\definecolor{paired-dark-pink}{RGB}{123, 65, 115}
\definecolor{paired-light-red}{RGB}{231, 150, 156}
\definecolor{paired-dark-red}{RGB}{131, 60, 56}
\definecolor{paired-light-yellow}{RGB}{231, 204, 149}
\definecolor{paired-dark-yellow}{RGB}{141, 109, 49}

\tikzset{%
    parent/.style =          {align=center,text width=2.8cm,rounded corners=3pt, line width=0.3mm, fill=gray!10,draw=gray!80},
    child/.style =           {align=center,text width=2.3cm,rounded corners=3pt, fill=blue!10,draw=blue!80,line width=0.3mm},
    grandchild/.style =      {align=center,text width=2cm,rounded corners=3pt},
    greatgrandchild/.style = {align=center,text width=1.5cm,rounded corners=3pt},
    greatgrandchild2/.style = {align=center,text width=1.5cm,rounded corners=3pt},
    referenceblock/.style =  {align=center,text width=1.5cm,rounded corners=2pt},
    acquisition/.style =     {align=center,text width=2.6cm,rounded corners=3pt, fill=paired-light-blue!50,draw=paired-dark-blue!65,line width=0.3mm},
    acquisition_work/.style = {align=center, text width=10cm,rounded corners=3pt, fill=paired-light-blue!50,draw=blue!0,line width=0.3mm},
    representation/.style =  {align=center,text width=2.6cm,rounded corners=3pt, fill=paired-light-orange!50,draw=paired-dark-orange!65,line width=0.3mm},
    representation_work/.style = {align=center,text width=10cm,rounded corners=3pt, fill=paired-light-orange!50,draw=red!0,line width=0.3mm},
    representation_work_2/.style = {align=center,text width=8.7cm,rounded corners=3pt, fill=paired-light-orange!50,draw=red!0,line width=0.3mm},
    probing/.style =         {align=center,text width=2.6cm,rounded corners=3pt, fill=paired-light-green!50,draw=paired-dark-green!75,line width=0.3mm},
    probing_work/.style =    {align=center,text width=10cm,rounded corners=3pt, fill=paired-light-green!50,draw=cyan!0,line width=0.3mm},
    cus_probing_work/.style = {align=center,text width=8.7cm,rounded corners=3pt, fill=paired-light-green!50,draw=cyan!0,line width=0.3mm},
    editing/.style =         {align=center,text width=2.2cm,rounded corners=3pt, fill=paired-light-purple!50,draw=paired-dark-purple!75,line width=0.3mm},
    editing_work/.style =    {align=center,text width=6cm,rounded corners=3pt, fill=paired-light-purple!50,draw=orange!0,line width=0.3mm},
    application/.style =     {align=center,text width=2.2cm,rounded corners=3pt, fill=paired-light-red!35,draw=paired-light-red!90,line width=0.3mm},
    application_work/.style = {align=center,text width=6cm,rounded corners=3pt, fill=paired-light-red!35,draw=magenta!0,line width=0.3mm},
}

\title{The Road to Artificial SuperIntelligence: \\A Comprehensive Survey of Superalignment}

\author{
HyunJin Kim\textsuperscript{1,2},
DongHyun Ryu\textsuperscript{2},
Xiaoyuan Yi\textsuperscript{1}\thanks{Corresponding authors: Xiaoyuan Yi and JinYeong Bak},
Jing Yao\textsuperscript{1},
Jianxun Lian\textsuperscript{1}\\
\textbf{Muhua~Huang\textsuperscript{3},
Shitong~Duan\textsuperscript{4},
JinYeong Bak\textsuperscript{2}$^*$, Xing~Xie\textsuperscript{1}} \\
\textsuperscript{1}Microsoft Research Asia, \textsuperscript{2}Sungkyunkwan University\\
\textsuperscript{3}Stanford University, \textsuperscript{4}Fudan University\\
\texttt{\url{khyunjin1993@skku.edu}},
\texttt{\url{xiaoyuanyi@microsoft.com}}, \texttt{\url{jy.bak@skku.edu}}
}

\begin{document}

\ifcolmsubmission
\linenumbers
\fi

\maketitle

\begin{abstract}
The emergence of large language models (LLMs) has sparked discussion on Artificial Superintelligence (ASI), a hypothetical AI system that surpasses human intelligence. Although ASI remains hypothetical and far beyond current AI capabilities, discussing its potential and exploring its feasibility and potential risks is critical for the development of future AI systems. The idea of superalignment originates from scalable oversight, which studies how to supervise increasingly capable AI systems when direct human supervision becomes insufficient. In this paper, we focus on the superalignment problem: ``The process of supervising, controlling, and governing artificial superintelligence.'' We first review scalable oversight paradigms---Sandwiching, Self-Enhancement, and Weak-to-Strong Generalization---then analyze the limitations of current paradigms through the lens of possibility and impossibility, discuss key challenges, and propose pathways for the safe and continual improvement of future AI systems.
\end{abstract}

\section{Introduction} \label{intro}
\begin{wrapfigure}{rt}{0.55\columnwidth}
    \centering
    \vspace{-0.5cm}
    \includegraphics[width=1.0\linewidth]{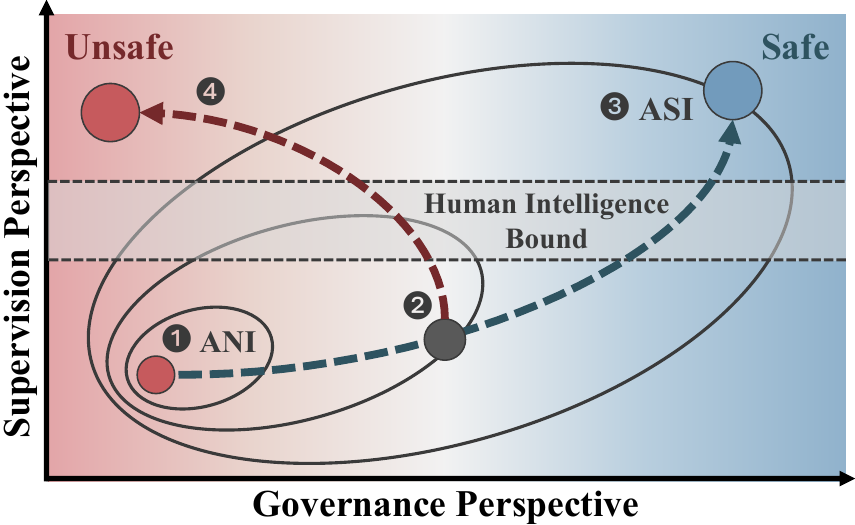}
    \caption{Superalignment challenge from the supervision and governance. \protect\BC{1} Current AI systems are in the ANI regime. \protect\BC{2} Progress toward potential AGI is possible under human supervision. \protect\BC{3} Superalignment aims to ensure capable AI systems beyond human intelligence and safe. \protect\BC{4} Failure to align such systems could pose potential risks.
    }
    \vspace{-0.5cm}
    \label{fig:motivation}
\end{wrapfigure}
Language models have shown improved capabilities as their scale increases, a phenomenon known as emergent abilities~\citep{NEURIPS2020_1457c0d6, kaplan2020scalinglawsneurallanguage, wei2022emergent, nam2024an, hu2024predicting} across a wide range of tasks~\citep{NEURIPS2020_1457c0d6, chen2021evaluatinglargelanguagemodels, shi2022languagemodelsmultilingualchainofthought, wei2022chain, rein2024gpqa, evalperf}. This scaling has led to the development of large language models (LLMs), including proprietary models such as GPT-5.4~\citep{openai2026gpt54}, Claude 4~\citep{claude45}, and Gemini 3.1 Pro~\citep{google2026gemini31}, as well as open-source models like Llama 4~\citep{llama4}, and Qwen 3.5~\citep{qwen35}.

Despite these advances, current AI systems are generally considered examples of Artificial Narrow Intelligence (ANI)~\citep{ibm-artificial-intelligence-types}, which refers to systems designed to perform specialized tasks, such as healthcare~\citep{xiao-etal-2025-online} or software engineering~\citep{hou2024largelanguagemodelssoftware}, but lack the broad cognitive abilities of humans. As LLMs' capabilities continue to improve, discussion on the possibility of Artificial General Intelligence (AGI) has emerged, referring to AI systems capable of human-level reasoning and adaptability across diverse domains~\citep{morrisposition23, hendrycks2025definitionagi, 2026_humanity}, which has led to increasing discussion about whether AGI could eventually be realized~\citep{bubeck2023sparks}.

Artificial Superintelligence (ASI) is a hypothetical concept beyond AGI in which AI systems surpass human intelligence across all domains~\citep{pohl2015artificial, batin2017artificial, ibm-superintelligence, pmlr-v235-hughes24a}. 
Considering the rapidly increasing capabilities of AI systems, it becomes increasingly important to anticipate the possibility of ASI and explore the challenges that may arise once AI systems surpass human oversight capabilities, particularly the lack of reliable supervision signals.

To address these challenges, the concept of \textit{superalignment} has been proposed, defined as ``the process of supervising, controlling, and governing artificial superintelligence systems''~\citep{openai-superalignment,ibm-superalignment}. Superalignment encompasses two key objectives from two complementary perspectives: (1) from a \textit{supervision perspective}, ensuring effective supervision of highly capable AI systems, and (2) from a \textit{governance perspective}, governing these systems so that their behavior remains safe and aligned with human values. Fig.~\ref{fig:motivation} illustrates the challenges of superalignment from a supervision and governance perspective.
Achieving both goals simultaneously is crucial for realizing safe and beneficial ASI~\citep{ibm-superintelligence}, as they cannot be separately achieved in practice.

In this paper, we aim to provide an overview of the superalignment problem through the lens of possibility and impossibility within existing paradigms. 
In Sec.~\ref{bg}, we first highlight the scope and focus
of this work, introduce the concepts of each AI system (\textit{i.e.,} ANI, AGI, and ASI), and present a formalization of the superalignment problem along with the associated challenges. 
In Sec.~\ref{approaches}, we analyze current scalable oversight paradigms that aim to supervise increasingly capable AI systems when direct human supervision becomes insufficient, discuss the specific challenges associated with each paradigm, and examine what is currently feasible and what remains unsolved. 
Finally, in Sec.~\ref{future_direction}, we summarize the key challenges across scalable oversight paradigms and explore potential pathways.
\section{Scope and Background} \label{bg}

\textbf{Our Scope.} While ASI remains far from realization, in this survey we analyze recent results on scalable oversight paradigms for achieving superalignment. We examine each existing paradigm from two perspectives to clarify its strengths and limitations: \textit{the possibilities} and \textit{the impossibilities of achieving superalignment}. We detail each of these perspectives as follows.

\noindent\textit{Possibility of Superalignment}. We analyze the conditions under which scalable oversight paradigms may enable superalignment. From the supervision perspective, we examine scalable oversight paradigms that allow weaker (\textit{i.e.,} noisy or insufficient) supervision signals to train more capable AI systems \citep{burns2023weaktostronggeneralizationelicitingstrong}. From the governance perspective, we examine whether these paradigms can steer increasingly capable systems so that their behavior remains consistent with human values and remains safe~\citep{ibm-superintelligence}.

\noindent\textit{Impossibility of Superalignment}. We analyze the limitations that may prevent scalable oversight paradigms from achieving superalignment. From the supervision perspective, we discuss challenges in generating reliable supervision signals for systems whose capabilities exceed human understanding (\textit{e.g.,} error propagation, reward hacking, and deceptive behaviors \citep{amodei2016concreteproblemsaisafety, hubinger2019risks, everitt2023causal}). From the governance perspective, we examine the difficulty of ensuring that increasingly capable AI systems remain aligned with human values and safe, particularly when such AI systems may exploit alignment mechanisms \citep{hubinger2019risks, chen2025aideceptionrisksdynamics}.

\subsection{Terminology}
Let $\mathcal{A}$ denote an AI system and $\mathcal{H}$ a human. Given a task $\mathbf{x}$ from the task space $\mathbf{x}\in\mathcal{X}$, including extremely difficult problems (\textit{e.g.,} ``Finding a Formal Proof for the P vs NP Problem''), $\mathbf{y}$ a gold or desired solution when such a reference is available, and $\hat{\mathbf{y}} \in \mathcal{Y}(\mathbf{x})$ a candidate output. Let $\mathcal{T}(\mathbf{x}, \hat{\mathbf{y}})$ denote the gold target criterion of interest, such as correctness, safety, alignment, or a joint objective over these properties. In this survey, we formulate alignment and superalignment \textit{as problems of learning the preference between candidate outputs}. Although the number of $\mathbf{y}$ can be extended to comparisons among multiple candidate outputs, we define it as a pairwise comparison as the basic unit of supervision. Given two outputs from AI system $\mathcal{A}$ (\textit{e.g.,} $\hat{\mathbf{y}}_a \sim \mathcal{A}_a(\mathbf{x})$ and $\hat{\mathbf{y}}_b \sim \mathcal{A}_b(\mathbf{x})$), the central question is whether an overseer $\mathcal{O}$ can reliably determine which output better satisfies $\mathcal{T}$ (\textit{i.e.,} $\hat{\mathbf{y}}_a \succ_{\mathcal{O}}^{\mathbf{x}} \hat{\mathbf{y}}_b$ and $\mathcal{T}(\mathbf{x},\hat{\mathbf{y}}_a) > \mathcal{T}(\mathbf{x},\hat{\mathbf{y}}_b)$). Such comparisons serve as the basis of useful supervision signals. Under this view, the distinction between alignment and superalignment lies in whether this comparison process remains reliable as task difficulty and model capability increase.

\textbf{Artificial Narrow Intelligence (ANI)} refers to an AI system $\mathcal{A}_{\text{ANI}}$ that outperforms or is comparable to human on a specific task~\citep{silver2016mastering, ibm-superintelligence} while performing below human level on other tasks due to limited generalization capabilities. Formally, there exist tasks $i \neq j$ such that $\mathcal{T}(\mathbf{x}_i,\mathcal{A}_{\text{ANI}}(\mathbf{x}_i)) > \mathcal{T}(\mathbf{x}_i,\mathcal{H}(\mathbf{x}_i))$ while $\mathcal{T}(\mathbf{x}_j,\mathcal{A}_{\text{ANI}}(\mathbf{x}_j)) \leq \mathcal{T}(\mathbf{x}_j,\mathcal{H}(\mathbf{x}_j))$.
\textbf{Artificial General Intelligence (AGI)} is an AI system ($\mathcal{A}_{\text{AGI}}$) beyond $\mathcal{A}_{\text{ANI}}$ that possesses human-level intelligence and demonstrates general-purpose capabilities across a wide range of tasks and domains~\citep{goertzel2014artificial, pohl2015artificial, batin2017artificial, fei2022towards, bubeck2023sparks}. Formally, $\mathcal{T}(\mathbf{x},\mathcal{A}_{\text{AGI}}(\mathbf{x})) \ge \mathcal{T}(\mathbf{x},\mathcal{H}(\mathbf{x}))$ for most $\mathbf{x} \in \mathcal{X}$. \textbf{Artificial Superintelligence (ASI)} is a hypothetical AI system ($\mathcal{A}_{\text{ASI}}$) that vastly surpasses human capabilities across all tasks $\mathcal{X}$ requiring cognitive reasoning~\citep{nick2014superintelligence, drexler2019reframing, ibm-superintelligence, morrisposition23, pmlr-v235-hughes24a}. Formally, $\mathcal{T}(\mathbf{x},\mathcal{A}_{\text{ASI}}(\mathbf{x})) \gg \mathcal{T}(\mathbf{x},\mathcal{H}(\mathbf{x}))$ for most tasks $\mathbf{x} \in \mathcal{X}$.
Although ASI remains far from realization, proactively addressing the challenges of superalignment is both timely for the continual improvement of task competence and for mitigating potential catastrophic risks~\citep{pueyo2018growth, hubinger2019risks}. Consequently, supervision and governance should co-evolve, simultaneously advancing AI capabilities while mitigating associated risks.

\subsection{Definition of Alignment}
\textbf{Existing alignment} trains an AI system so that its outputs are preferred by a human overseer and better satisfy $\mathcal{T}$. Formally, given a task $\mathbf{x}$ and two candidate outputs $\hat{\mathbf{y}}_a,\hat{\mathbf{y}}_b$, existing alignment assumes that the human overseer $\mathcal{H}$ can provide a reliable comparison $\hat{\mathbf{y}}_a \succ_{\mathcal{H}}^{\mathbf{x}} \hat{\mathbf{y}}_b$ when $\mathcal{T}(\mathbf{x},\hat{\mathbf{y}}_a) > \mathcal{T}(\mathbf{x},\hat{\mathbf{y}}_b)$. Thus, \textit{existing alignment operates in regimes where the task remains within the evaluative capability of the human overseer $\mathcal{H}$}, so that comparisons remain reliable.

Current alignment paradigms, such as reinforcement learning with human feedback (RLHF)~\citep{christiano2017deep, ouyang2022training, gulcehre2023reinforcedselftrainingrestlanguage}, supervised fine-tuning (SFT)~\citep{zhou2023lima, rafailov2023direct}, and in-context learning (ICL)~\citep{xu2023align, gou2024critic}, have shown promise in aligning LLMs with human values and goals~\citep{ijcai2024p0918}. However, these alignment methods pose challenges.
\textit{From a supervision perspective}, current alignment paradigms are limited by the scalability and reliability of human feedback \citep{bowman2022measuringprogressscalableoversight, saunders2022selfcritiquingmodelsassistinghuman}. As model capabilities increase, obtaining accurate labeled data becomes increasingly difficult \citep{rein2024gpqa}, particularly for supervising a potential $\mathcal{A}_{\text{ASI}} \citep{burns2023weaktostronggeneralizationelicitingstrong}$.
\textit{From a governance perspective}, increasing model capability also raises the risk of alignment faking or deceptive behavior~\citep{hubinger2019risks, hubinger2024sleeperagentstrainingdeceptive, yang2024superficialalignmentstrongmodelsdeceive}, which further complicates the governance of $\mathcal{A}_{\text{ASI}}$.

\textbf{Superalignment} arises when humans can no longer reliably evaluate outputs with respect to $\mathcal{T}$. To express this, for a given task $\mathbf{x}$, we define $\hat{\mathbf{y}}_a \succ_{\mathcal{O}}^{\mathbf{x}} \hat{\mathbf{y}}_b$ to mean that an overseer $\mathcal{O}$ (\textit{i.e.}, both $\mathcal{A}$ and $\mathcal{H}$) cannot provide direct and reliable supervision that $\hat{\mathbf{y}}_a$ is better than $\hat{\mathbf{y}}_b$ on task $\mathbf{x}$, even when $\mathcal{T}(\mathbf{x},\hat{\mathbf{y}}_a) > \mathcal{T}(\mathbf{x},\hat{\mathbf{y}}_b)$. In other words, \textit{the overseer can no longer directly distinguish better outputs from worse ones}.

The goal of superalignment is to increase capabilities while aligning ASI~\citep{openai-superalignment, ibm-superalignment}. Superalignment is defined as ``the process of supervising, controlling, and governing artificial superintelligence systems''~\citep{ibm-superalignment}, and aims to develop methods that provide high-quality supervision signals, particularly in the absence of human oversight. The concept originates from scalable oversight, formally defined as ``the process of ensuring that a given AI system adheres to aspects of its objectives that are too costly or impractical to evaluate frequently during training''~\citep{amodei2016concreteproblemsaisafety}. This perspective highlights the challenge of guiding AI systems with supervision that scales beyond human capabilities, where obtaining reliable, high-quality supervision signals becomes increasingly costly or even infeasible as tasks grow more complex~\citep{leike2018scalable, christiano2018ai, pang2022qualityquestionansweringlong}.
The emergence of increasingly capable models led attention to the superalignment problem as a way to ensure that AI systems remain aligned with human values as their capabilities grow~\citep{openai-superalignment}. Scalable oversight paradigms~\citep{kim2026researchsuperalignmentadvancealternating}, such as sandwiching~\citep{cotra} and weak-to-strong generalization (W2SG)~\citep{burns2023weaktostronggeneralizationelicitingstrong}, have emerged as promising approaches.
\section{Approaches and Methods} \label{approaches}
Several scalable oversight techniques have been proposed, including iterated distillation and amplification (IDA)~\citep{christiano2018ai, IDAweb}, recursive reward modeling (RRM)~\citep{leike2018scalable}, and cooperative inverse reinforcement learning~\citep{NIPS2016_c3395dd4}. These approaches form the foundation of scalable oversight methods, such as sandwiching (Human$\leftrightarrow$AI), self-enhancement (AI$\leftrightarrow$AI), and weak-to-strong generalization (Weak AI/Data$\rightarrow$Strong AI)~\citep{irving2018debate, bai2022constitutionalaiharmlessnessai, bowman2022measuringprogressscalableoversight, burns2023weaktostronggeneralizationelicitingstrong}. Fig.~\ref{fig:updated_taxonomy} and~\ref{fig:paradigm_overview} illustrates the landscape of scalable oversight paradigms.

\subsection{Sandwiching}
\label{sandwiching}

Sandwiching~\citep{christiano2018ai,leike2018scalable,irving2018debate,cotra,bowman2022measuringprogressscalableoversight} studies how collaboration between humans ($\mathcal{H}$) and AI systems ($\mathcal{A}$) can produce an answer $\hat{\mathbf{y}}$ for an input $\mathbf{x}$ even when neither can reliably solve the task alone. Its core intuition is that AI systems may offer broader knowledge, faster search, or stronger pattern recognition, while humans are better able to evaluate, decompose, question, or verify candidate answers. In this setting, the overseer is typically a human $\mathcal{H}$ who evaluates outputs $\hat{\mathbf{y}} \sim \mathcal{A}(\mathbf{x})$, where $\mathcal{A}$ may denote a single AI or multiple AI systems $(\mathcal{A}_1,\cdots,\mathcal{A}_K)$. Although the human may not reliably evaluate a final output, structuring interaction around intermediate evidence or arguments or subproblem solutions enables more reliable comparisons that remain informative with respect to $\mathcal{T}$.

Sandwiching can be separated into two distinct concepts. \textbf{Debate} uses two AI systems $\mathcal{A}_a$ and $\mathcal{A}_b$, which are given candidate answers $\hat{\mathbf{y}}_a$ and $\hat{\mathbf{y}}_b$~\citep{irving2018debate}. Each generates an argument $\mathbf{g}_a$ and $\mathbf{g}_b$ to support its assigned answer (\textit{i.e.,} $\mathcal{A}_a$ argues that $\hat{\mathbf{y}}_a \succ_{\mathcal{A}_a}^{\mathbf{x}} \hat{\mathbf{y}}_b$, while $\mathcal{A}_b$ argues that $\hat{\mathbf{y}}_b \succ_{\mathcal{A}_b}^{\mathbf{x}} \hat{\mathbf{y}}_a$). The overseer $\mathcal{O}$ (\textit{i.e.,} human) then compares the arguments (\textit{e.g.,} $\mathbf{g}_a \succ_{\mathcal{O}}^{\mathbf{x}} \mathbf{g}_b$) and selects the final prediction $\hat{\mathbf{y}}$. Debate succeeds when such comparisons induced by the arguments remain faithful to the target criterion, \textit{i.e.,} when $\mathbf{g}_a \succ_{\mathcal{O}}^{\mathbf{x}} \mathbf{g}_b$ implies $\mathcal{T}(\mathbf{x}, \hat{\mathbf{y}}_a) > \mathcal{T}(\mathbf{x}, \hat{\mathbf{y}}_b)$. \textbf{Task Decomposition and Recursive Supervision} involve an overseer $\mathcal{O}$ that decomposes $\mathbf{x}$ into subtasks $(x_1,\dots,x_K)=\mathcal{O}(\mathbf{x})$~\citep{christiano2018ai}. An AI system then produces subanswers $\hat{y}_i \sim \mathcal{A}(x_i)$, which are evaluated and aggregated by the overseer to obtain the final prediction $\hat{\mathbf{y}}=\mathcal{O}(\hat{y}_1,\dots,\hat{y}_K)$. This paradigm succeeds when the overseer can reliably evaluate the intermediate subanswers $(\hat{y}_a^1,\dots,\hat{y}_a^K)$ compared to $(\hat{y}_b^1,\dots,\hat{y}_b^K)$ (\textit{e.g.,} $\hat{y}^1_a \succ_{\mathcal{O}}^{\mathbf{x}} \hat{y}^1_b$ implies $\mathcal{T}(\mathbf{x}, \hat{y}^1_a) > \mathcal{T}(\mathbf{x}, \hat{y}^1_b)$), and when these evaluations compose into improvement with respect to the overall target criterion $\mathcal{T}(\mathbf{x}, \hat{\mathbf{y}}_a) > \mathcal{T}(\mathbf{x}, \hat{\mathbf{y}}_b)$.

\subsubsection{Core Concept \& Analysis}
Early discussions of sandwiching as a scalable oversight paradigm led to the following related concepts. \textbf{Debate.}~\cite{irving2018debate} proposes adversarial debate, where two AI systems $\mathcal{A}_a, \mathcal{A}_b$ present competing arguments and a weaker judge (overseer) $\mathcal{O}$ selects the more truthful one, based on the idea that evaluation can be easier than generation. 
In parallel, \textbf{Task Decomposition and Recursive Supervision.} Iterated amplification~\citep{christiano2018ai} proposes recursively decomposing complex tasks into simpler subproblems $\mathbf{x}=(x_1, \dots, x_K)$ that can be solved by a weaker judge $\mathcal{O}$. \cite{leike2018scalable} extends this approach through recursive reward modeling, learning reward signals from decomposed human judgments rather than direct evaluation of difficult tasks. 

Building on these concept, \cite{cotra} initialized an early discussion and \cite{bowman2022measuringprogressscalableoversight} operationalized sandwiching as an empirical paradigm by focusing on tasks where \emph{human specialists succeed, unaided non-expert humans fail, and current AI systems also fail}, enabling the measurement of scalable oversight. They further show that AI-assisted non-experts ($\hat{\mathbf{y}}_{\mathcal{H},\mathcal{A}} \sim \mathcal{O}(\mathbf{x})$) can outperform both unaided humans ($\hat{\mathbf{y}}_{\mathcal{H}} \sim \mathcal{H}(\mathbf{x})$) and standalone AI systems ($\hat{\mathbf{y}}_{\mathcal{A}} \sim \mathcal{A}(\mathbf{x})$), providing a proof of concept for human--AI oversight on sufficiently difficult tasks (\textit{e.g.,} \cite{pang2022qualityquestionansweringlong}).
Subsequent work analyzes when such interaction protocols are effective. \cite{kenton2024scalableoversightweakllms} finds that debate is most beneficial in settings with information asymmetry. Similarly, \cite{khan2024debatingpersuasivellmsleads} shows that sandwiching is more effective under debate than consultancy, and that optimizing for persuasiveness improves truthfulness in debate. From a theoretical perspective, \cite{browncohen2023scalableaisafetydoublyefficient} introduces doubly-efficient debate, where a truthful debater can generate correct arguments efficiently, while a limited judge can check them using only a small amount of computation. \cite{young2026knowledgedivergencevaluedebate} further finds that debate is most effective when AI systems have diverse knowledge.

\subsubsection{Improvement \& Optimization Strategy}
Building on human--AI collaboration~\citep{10.1145/3449287,10.1145/3531146.3533193,Bondi_Koster_Sheahan_Chadwick_Bachrach_Cemgil_Paquet_Dvijotham_2022,ijcai2022p341} and early conceptual discussions~\citep{irving2018debate,christiano2018ai,leike2018scalable,cotra,bowman2022measuringprogressscalableoversight}, several methods are proposed.

\textbf{Debate \& Multi-agent Oversight.}
\cite{kirchner2024proververifiergamesimprovelegibility} proposes legibility training against small verifiers (AI systems) to improve the human verifiability of outputs from capable AI systems. Outside the core scalable oversight setting but within multi-agent debate, \cite{sun-etal-2025-cortexdebate} identifies limitations of standard debate (\textit{i.e.,} long debate contexts) and proposes sparse debating. \cite{shigemura2025recursiveknowledgesynthesismultillm} explores tri-agent cross-validation to improve stability and interpretability in a multi-agent debate setup.
\textbf{Task Decomposition and Recursive Supervision.}
\cite{prasad-etal-2024-adapt} proposes adaptive task decomposition (ADaPT) to better handle varying task complexity and model capability. However, hierarchical reasoning methods still face challenges such as weakened continuity and increased computational overhead. ReCAP~\citep{zhang2025recap} proposes a hierarchical reasoning framework with shared context to mitigate fragmentation and inefficiency. \cite{wen2026scalableoversightsuperhumanai} studies recursive critique as a scalable supervision signal, hypothesizing that critique of critique can be easier than critique itself and that this relationship may hold recursively.

\paragraph{Application.} Sandwiching has been explored in diverse domains. \textbf{Error Detection}~\citep{chen2026towards} applies collaborative multi-agent debate to identify errors. \textbf{Competitive Programming}~\citep{wen2025learningtaskdecompositionassist} studies task decomposition to assist human oversight by breaking complex solutions into simpler subtasks. \textbf{Code Evaluation}, \citep{mcaleese2024llmcriticshelpcatch} trains LLM critics to help humans assess model-written code. \textbf{Cultural Alignment}~\citep{ki-etal-2025-multiple} proposes a multi-agent debate framework with culturally grounded scenarios. More broadly, aside from scalable oversight, debate is used to improve factuality and reasoning~\citep{du2023improvingfactualityreasoninglanguage}, explore ethically nuanced question answering~\citep{sturgeoninvestigating}, and evaluate whether LLMs can be trusted as evaluators~\citep{chern2024largelanguagemodelstrusted}.

\subsubsection{Possibility \& Impossibility}
\paragraph{The Possibility.}
The possibility of sandwiching lies in converting tasks that are too difficult for humans to evaluate end-to-end into sequences of easier local judgments over decomposed subproblems. \textbf{Increasing human judge confidence.} This becomes more plausible when sandwiching methods expose multiple plausible answers~\citep{10.5555/3692070.3692445} or request additional information to resolve uncertainty~\citep{testoni-fernandez-2024-asking}, enabling more reliable human judgements. \textbf{Reducing Reliance on Human Labeling.} Efficient supervision requires selecting the appropriate judge (human or AI). Learning-to-defer (L2D) provides a principled framework for optimal human--AI decision routing~\citep{10.1609/aaai.v37i5.25742,mozannar2023predictexactalgorithmslearning} and efficient allocation of human judgments~\citep{mohankumar-khapra-2022-active}. \textbf{Information Asymmetry.} Sandwiching is most promising when humans and AI possess complementary information or capabilities, and when tasks are decomposed into subtasks within the judge's competence~\citep{10.1145/3534561, hemmer2024complementarityhumanaicollaborationconcept}.

\paragraph{The Impossibility.}
The main limitation of sandwiching is that human judgment is inherently fragile. \textbf{Biased Evaluator.} Human judgments can be biased by persuasive but incorrect explanations~\citep{10.1145/3397481.3450639}, especially as reliance on AI increases~\citep{10.1145/3579605}. Decision-making is also sensitive to self-confidence, which can affect AI-assisted outcomes~\citep{10.1145/3613904.3642671}. \textbf{Limited Real-world Deployment.} Optimal human--AI decision routing may be impractical due to L2D requirements, such as collecting per-instance predictions from all human decision-makers, which is often infeasible~\citep{leitao2022humanaicollaborationdecisionmakinglearning,10.1609/aaai.v37i5.25742}. \textbf{Sycophancy.} Both humans and AI systems may favor fluent but incorrect answers~\citep{sharma2025understandingsycophancylanguagemodels}. \textbf{Deception.} Capable AI systems may strategically manipulate judges rather than expose useful information for oversight~\citep{Hagendorff_2024}.

\subsection{Self-Enhancement}
\label{constitutional_ai_rlaif}
Self-Enhancement~\citep{bai2022constitutionalaiharmlessnessai,huang2024far} builds on self-play~\citep{silver2017masteringchessshogiselfplay} and Expert Iteration~\citep{anthony2017thinkingfastslowdeep}, using feedback from an AI system $\mathcal{A}$ itself or from multiple AI systems $\{\mathcal{A}_1, \ldots, \mathcal{A}_K\}$ to generate a solution $\hat{\mathbf{y}}$ for an input $\mathbf{x}$ with minimal or no human supervision (\textit{i.e.,} $\mathcal{O}=\mathcal{A}$). The key intuition in self-enhancement is that outputs can be iteratively refined over multiple rounds $i$. Formally, given an input $\mathbf{x}$, an AI system $\mathcal{A}_{i}$ at iteration $i$ produces a candidate output $\hat{\mathbf{y}}_{i+1} \sim \mathcal{A}_{i}(\mathbf{x})$, which is then used to improve the next-iteration model $\mathcal{A}_{i+1}$. The key assumption is that AI-generated oversight remains informative enough for iterative improvement, \textit{e.g.,} an AI overseer can reliably compare candidate outputs such that $\hat{\mathbf{y}}_{i+1} \succ_{\mathcal{A}_{i}}^{\mathbf{x}} \hat{\mathbf{y}}_{i}$ implies $\mathcal{T}(\mathbf{x},\hat{\mathbf{y}}_{i+1}) > \mathcal{T}(\mathbf{x},\hat{\mathbf{y}}_{i})$.

\subsubsection{Core Concept \& Analysis}
Early work studies whether models can identify and correct errors in their own outputs. \cite{saunders2022selfcritiquingmodelsassistinghuman} trains models to generate critiques that help detect errors, showing that larger models produce more useful critiques and can leverage them to improve their outputs. Similarly, \cite{ganguli2023capacitymoralselfcorrectionlarge} show that sufficiently capable models can avoid harmful responses when prompted to reflect on their own outputs.
Building on this idea, subsequent work explores frameworks that use AI-generated feedback for scalable supervision, differing in how and where the signal is applied. \textbf{In-Context Refinement (ICR)} improves outputs at inference time without updating model parameters~\citep{madaan2023selfrefine,wang2023selfconsistency,lin2024the,gou2024critic,cai2026codecontestsopoweringllmsfeedbackdriven}. \textbf{Learning with AI Feedback (LAIF)} incorporates model-generated feedback into training~\citep{zelikman2022starbootstrappingreasoningreasoning,gulcehre2023reinforcedselftrainingrestlanguage,ji2024aligner,yuan2024selfrewardinglanguagemodels,10.5555/3692070.3693141,lai2024stepdpostepwisepreferenceoptimization,kumar2024traininglanguagemodelsselfcorrect,sharma2024a,li2025curriculumrlaifcurriculumalignmentreinforcement,yu-etal-2025-diverse,fei2025postcompletionlearninglanguagemodels}. \textbf{Multi-agent Oversight} leverages interactions among multiple AI systems to build a supervision signal~\citep{wang2023selfconsistency,luo2023critiqueabilitylargelanguage,chern2024largelanguagemodelstrusted,pang2024selfalignment,Negozio_2025}.

Alongside these developments, other work also analyzes the reliability and limitations of self-enhancement. \cite{luo2023critiqueabilitylargelanguage} shows that critique ability improves with scale, but effective self-critique remains challenging. \cite{chern2024largelanguagemodelstrusted} studies whether models can act as reliable evaluators (\textit{i.e.,} overseers) and proposes a meta-evaluation framework based on debate. Other work explores the co-evolution of capability and safety~\citep{lab2025safeworkr1coevolvingsafetyintelligence}, open-ended self-improvement via self-modification~\citep{zhang2025darwingodelmachineopenended}, and theoretical guarantees for iterative self-rewarding optimization~\citep{fu2026selfrewardingworkstheoreticalguarantees}.

\subsubsection{Improvement \& Optimization Strategy}
\textbf{In-Context Refinement.}
It improves outputs at inference time without parameter updates by leveraging self-generated feedback and reasoning. \citep{madaan2023selfrefine} proposes iterative self-refinement via model-generated critiques, while \cite{wang2023selfconsistency} improves reasoning through diverse reasoning path sampling. \cite{gou2024critic} introduces tool-assisted refinement, and \cite{lin2024the} shows that alignment behaviors can be elicited through a few well-written in-context demonstrations. \cite{cai2026codecontestsopoweringllmsfeedbackdriven} further refines test cases using execution-based feedback.
\textbf{Learning with AI Feedback.}
A large body of work leverages AI-generated feedback to improve alignment and capability without relying on human supervision. Early approaches bootstrap reasoning or training data~\citep{zelikman2022starbootstrappingreasoningreasoning,gulcehre2023reinforcedselftrainingrestlanguage}, while later methods use synthetic preferences and reward signals for iterative optimization~\citep{ji2024aligner,yuan2024selfrewardinglanguagemodels,10.5555/3692070.3693141}. Subsequent work studies data quality and training dynamics, including the effectiveness of self-generated data~\citep{lai2024stepdpostepwisepreferenceoptimization,kumar2024traininglanguagemodelsselfcorrect,sharma2024a}, curriculum-based preference construction~\citep{li2025curriculumrlaifcurriculumalignmentreinforcement}, diversified feedback~\citep{yu-etal-2025-diverse}, and post-completion learning for joint reasoning and evaluation~\citep{fei2025postcompletionlearninglanguagemodels}.
\textbf{Multi-agent Enhancement.} 
are often combined with ICR and LAIF to improve supervision quality~\citep{wang2023selfconsistency,luo2023critiqueabilitylargelanguage,chern2024largelanguagemodelstrusted,pang2024selfalignment,Negozio_2025}. For example, \cite{pang2024selfalignment} explores collective self-improvement via multi-agent simulation, while \cite{Negozio_2025} proposes mutual verification of alignment proofs across isolated AI systems.

\paragraph{Application.} Recent work extends self-enhancement to diverse modalities. \textbf{Multimodal Reasoning.} \citep{yu2024rlaifvaligningmllmsopensource, ji-lu-2025-reflair} extend these approaches to multimodal settings, with RLAIF-V achieving strong performance on hallucination benchmarks and ReFLAIR enabling structured reflection via hybrid reward learning. \textbf{Code Generation.} \cite{dutta-etal-2024-applying} applies RLAIF to code generation, showing that AI feedback from larger models improves performance over fine-tuned models. \textbf{Medical Report Generation.} \cite{xiao-etal-2025-online} proposes an online self-alignment framework for radiology report generation using iterative self-generated data and evaluation.

\subsubsection{Possibility \& Impossibility}
\paragraph{The Possibility.} AI-generated supervision remains sufficiently informative to improve capability and alignment without human involvement~\citep{10.5555/3692070.3693141,yuan2024selfrewardinglanguagemodels}. \textbf{Successful Iterative Refinement.} Iterative refinement can improve outputs over time (\textit{i.e.,} $\mathcal{T}(\mathbf{x},\hat{\mathbf{y}}_{i+1}) > \mathcal{T}(\mathbf{x},\hat{\mathbf{y}}_{i})$)~\citep{madaan2023selfrefine,wang2023selfconsistency}. \textbf{Diverse Supervision.} Multiple AI systems can provide complementary supervision signals that are more informative than those from a single system~\citep{pang2024selfalignment}.

\paragraph{The Impossibility.} Several failure modes of self-enhancement may exist. \textbf{Distributional Collapse.} Recursive training on self-generated data can reduce diversity by concentrating on the model's own distribution~\citep{feng-etal-2023-dunst,shumailov2024curserecursiontraininggenerated}. \textbf{Self-preference Bias.} Models may prefer outputs aligned with their own generation distribution rather than selecting objectively better responses~\citep{wataoka2024selfpreference}. \textbf{Error Accumulation.} Errors can propagate across iterations, reducing diversity or inducing polarization~\citep{du2023minimizingaccumulatedtrajectoryerror,10.5555/3692070.3693106,piao2025emergencehumanlikepolarizationlarge}. \textbf{Deceptive Behavior.} Deceptive or misaligned behaviors may persist, creating a false impression of alignment~\citep{hubinger2024sleeperagentstrainingdeceptive}. \textbf{Forgetting.} Iterative training can degrade previously acquired capabilities or alignment during later optimization stages~\citep{fernando2025understandingforgettingllmsupervised}.

\subsection{Weak-to-Strong Generalization (W2SG)}
\label{w2sg}
W2SG~\citep{burns2023weaktostronggeneralizationelicitingstrong} studies whether weaker supervision (\textit{e.g.,} from noisy or easier data) can be used to train a stronger AI system. Let $(\mathcal{A}_1,\cdots,\mathcal{A}_K)$ denote AI systems ordered from weaker to stronger capability. A weaker model $\mathcal{A}_{i-1}$ generates supervision data $\mathcal{D}_{i-1}=\{(\mathbf{x},\hat{\mathbf{y}}_{i-1})\}$ with $\hat{\mathbf{y}}_{i-1} \sim \mathcal{A}_{i-1}(\mathbf{x})$, which is then used to train or supervise a stronger model $\mathcal{A}_{i}$. The key assumption is that weaker or noisier supervision still preserves enough information about the target criterion $\mathcal{T}$ to guide a stronger model toward better behavior (\textit{i.e.,} training on $\mathcal{D}_{{i-1}}$ can improve $\mathcal{A}_{i}$ beyond the capability of $\mathcal{A}_i$ itself). In particular, $\mathcal{A}_{i}$ may compare its own outputs $\hat{\mathbf{y}}_{i} \sim \mathcal{A}_{i}(\mathbf{x})$ with supervision $\hat{\mathbf{y}}_{i-1}$ to refine its behavior. Thus, $\hat{\mathbf{y}}_{i} \succ_{\mathcal{A}_{i}}^{\mathbf{x}} \hat{\mathbf{y}}_{i-1}$ implies $\mathcal{T}(\mathbf{x},\hat{\mathbf{y}}_{i}) > \mathcal{T}(\mathbf{x},\hat{\mathbf{y}}_{i-1})$ or $\hat{\mathbf{y}}_{i-1} \succ_{\mathcal{A}_{i}}^{\mathbf{x}} \hat{\mathbf{y}}_{i}$ implies $\mathcal{T}(\mathbf{x},\hat{\mathbf{y}}_{i-1}) > \mathcal{T}(\mathbf{x},\hat{\mathbf{y}}_{i})$. The objective is to obtain supervision by leveraging the stronger model's latent knowledge~\citep{mallen2024elicitinglatentknowledgequirky}, even when the supervision from $\mathcal{A}_{i-1}$ is imperfect. W2SG builds on ideas from iterated amplification and recursive reward modeling~\citep{leike2018scalable,christiano2018ai,IDAweb}.

\subsubsection{Core Concept \& Analysis}
\textbf{Naive W2SG.} Early work on scalable oversight introduced theoretical frameworks~\citep{irving2018debate, christiano2018ai, leike2018scalable, hubinger2019risks, demski2020embeddedagency}. Building on this, \cite{burns2023weaktostronggeneralizationelicitingstrong} proposes W2SG, which explores the challenge of aligning future ASI systems through a simplified setup of human supervision of superhuman models, formalized through the question: ``Can weak models supervise stronger models?'' \textbf{Easy-to-Hard Generalization.} A related line of work studies supervision through reward models, showing that training on easier data can improve performance on harder tasks~\citep{sun2024easytohardgeneralizationscalablealignment}. This aligns with the broader intuition that learning from easier tasks can support generalization to more difficult ones~\citep{10.1145/237814.237833,hase-etal-2024-unreasonable}.

Subsequent work analyzes the mechanisms underlying W2SG. Theoretical studies identify conditions under which weak supervision remains informative, including pseudolabel correction~\citep{10.5555/3157382.3157497,10.5555/3524938.3525245}, coverage expansion~\citep{10.5555/2976040.2976052,pmlr-v139-cai21b,wei2022theoreticalanalysisselftrainingdeep}, and representation-based generalization effects~\citep{10.5555/3495724.3496382, charikar2024quantifying}. Related work further highlights compensation for teacher under-regularization, advantages of better-suited student regularization, and interactions between weak supervision and pretraining that enable the student to learn easy patterns from the teacher while retaining harder knowledge acquired during pretraining~\citep{66ace186-d33e-34c6-b587-c776d756007c,bai2010spectral,pmlr-v130-richards21b, moniri2025mechanismsweaktostronggeneralizationtheoretical}.
Recent work further refines these insights by relaxing assumptions and extending theoretical frameworks. These include generalizing misfit-based analyses~\citep{mulgund2025relating}, relaxing convexity assumptions~\citep{xu2025emergenceweaktostronggeneralizationbiasvariance}, studying alternative loss functions such as $f$-divergence~\citep{yao2025weaktostronggeneralizationfdivergence}, and analyzing W2SG in high-dimensional and representation-based regimes~\citep{ildiz2025highdimensionalanalysisknowledgedistillation,xue2025representationsshapeweaktostronggeneralization}.

\subsubsection{Improvement \& Optimization Strategy}
A growing line of work improves W2SG performance along three perspectives. (1) supervision construction, (2) data selection, and (3) iterative refinement. These categories reflect the primary focus of improvement and are not mutually exclusive.

\textbf{Supervision Construction.} Methods improve the quality of weak supervision through algorithmic and modeling advances, including data-efficient training~\citep{larsen2022optimalweakstronglearning}, ensemble and recursive supervision~\citep{sang2024improvingweaktostronggeneralizationscalable}, reliability-aware filtering and reweighting~\citep{guo2024improvingweaktostronggeneralizationreliabilityaware}, contrastive decoding~\citep{li-etal-2023-contrastive, jiang2025contrastiveweaktostronggeneralization}, and leveraging both successful and failed trajectories~\citep{ye2026weaktostronggeneralizationfailuretrajectories}.
\textbf{Data Selection.} Data-centric approaches focus on selecting or refining supervision signals, including overlap-based data selection~\citep{shin2025weaktostrong}, selective labeling with strong-model self-labeling~\citep{9710505, lang2025selectiveweaktostronggeneralization}, and joint refinement of weak labels and queries to mitigate overfitting~\citep{shi-etal-2025-mitigate}.
\textbf{Iterative Refinement.} Iterative methods improve supervision through repeated interaction between models, including co-supervised training with multiple teachers~\citep{liu2024cosupervisedlearningimprovingweaktostrong}, token-level ensemble refinement~\citep{agrawal2025ensemw2senhancingweaktostronggeneralization}, active teacher--student learning~\citep{wu2025aliceproactivelearningteachers}, and multi-agent contrastive preference optimization~\citep{lyu2025macpo}.

\paragraph{Application.}
\cite{yang2024weaktostrongreasoning} applies W2SG to complex reasoning tasks in AI systems through progressive learning. 
\citep{guo2024visionsuperalignmentweaktostronggeneralization} extends W2SG to vision foundation models using adaptive confidence distillation, allowing the student model to balance learning from weak supervision while relying on its own predictions when confident.

\subsubsection{Possibility \& Impossibility}

\paragraph{The Possibility.} A central question in W2SG is whether weak supervision can enable a stronger model to exceed its supervisor. \textbf{Eliciting Latent Capability.} Weak or noisy supervision can provide sufficient signal for a stronger model, eliciting latent capabilities present in the stronger model~\citep{burns2023weaktostronggeneralizationelicitingstrong}. \textbf{Theoretical Grounding.} Theoretical work identifies conditions under which weak supervision remains informative, including pseudolabel correction, coverage expansion, favorable representation structure, and appropriate regularization~\citep{lang2024theoreticalanalysisweaktostronggeneralization,charikar2024quantifying,moniri2025mechanismsweaktostronggeneralizationtheoretical}. \textbf{Appropriate Supervision Gap.} More recent analyses show that W2SG succeeds when weak supervision is sufficiently informative and the strong model can leverage prior knowledge from pre-training rather than merely imitating weak labels~\citep{mulgund2025relating,xu2025emergenceweaktostronggeneralizationbiasvariance,yao2025capabilitieslimitationsweaktostronggeneralization}.

\paragraph{The Impossibility.} This arises when weak supervision fails to provide reliable signals for training stronger systems. In such cases, \textbf{Weak Model Replication.} the strong model may inherit errors, overfit to noisy labels, or fail to generalize beyond the teacher's competence, limiting the effectiveness of W2SG~\citep{christiano2022formalizingpresumptionindependence, yao2025capabilitieslimitationsweaktostronggeneralization}. \textbf{Distribution Shift.} These limitations worsen under distribution shift, where weak supervision no longer matches the target distribution, potentially causing W2SG to underperform even the weak supervisor~\citep{jeon2025weaktostrong}. \textbf{Deceptive Behavior.} Moreover, W2SG may induce deceptive failure~\citep{yang2024superficialalignmentstrongmodelsdeceive}. \textbf{Reward Overoptimization.} In easy-to-hard generalization, overly optimizing against an imperfect reward model can lead to reward overoptimization and degraded ground-truth performance~\citep{10.5555/3618408.3618845}.
\section{Shared Bottlenecks across Paradigms and Future Directions} \label{future_direction}
Although scalable oversight paradigms differ in their supervision source and interaction method, they pursue one concrete goal \textit{``how to construct supervision signals that remain useful \& safe as model capability exceeds direct human oversight''}. In this sense, the paradigms differ only where the supervision comes from and under what conditions it remains informative. 

Supervision signal in scalable oversight must satisfy the following principles. \textbf{Informativeness Gap}: It must correlate with truth or aligned behavior strongly enough to guide improvement when neither direct human judgment nor weak supervision is fully reliable~\citep{bowman2022measuringprogressscalableoversight,burns2023weaktostronggeneralizationelicitingstrong,10.5555/3692070.3693141}. \textbf{Recursive Instability}: It must remain stable under recursive or iterative use, without compounding errors, collapsing diversity, or degrading into brittle heuristics that work only in simplified settings~\citep{feng-etal-2023-dunst,du2023minimizingaccumulatedtrajectoryerror,shumailov2024curserecursiontraininggenerated,fernando2025understandingforgettingllmsupervised}. \textbf{Strategic Vulnerability}: It must remain robust to strategic behavior, including deception, manipulation, reward hacking, or persuasive failure modes~\citep{hubinger2019risks,everitt2023causal,Hagendorff_2024,yang2024superficialalignmentstrongmodelsdeceive}. \textbf{Generalization}: It must generalize beyond controlled environments, narrow task classes, or simplifying assumptions, to realistic high-stakes domains~\citep{bowman2022measuringprogressscalableoversight,charikar2024quantifying,lang2024theoreticalanalysisweaktostronggeneralization,sharma2024a}. \textbf{Static Oversight Failure}: It must remain effective as the capability gap between the supervised system and its supervisor grows, since fixed supervision may become unreliable once the system substantially exceeds the overseer's competence~\citep{cotra,burns2023weaktostronggeneralizationelicitingstrong,10.1609/aaai.v37i5.25742,mozannar2023predictexactalgorithmslearning}. A comprehensive summary is in Appendix Tab.~\ref{tab:bottlenecks_directions}.

These observations show that \textit{future progress toward superalignment should move beyond paradigm-specific improvements and focus on shared principles for scalable oversight}. We specify potential directions: \textbf{Improve the Reliability} of imperfect supervision through better evaluation protocols, uncertainty estimation, and supervision verification, so that humans, weak models, and AI overseer can provide signals whose failure modes are measurable rather than opaque~\citep{ijcai2022p341,Bondi_Koster_Sheahan_Chadwick_Bachrach_Cemgil_Paquet_Dvijotham_2022, 10.1609/aaai.v37i5.25742, 10.5555/3692070.3692445}. \textbf{Preserve Quality Supervision} under recursive improvement, preventing error accumulation, collapse, forgetting, and other self-reinforcing pathologies across multiple refinement or training cycles~\citep{feng-etal-2023-dunst,shumailov2024curserecursiontraininggenerated,kumar2024traininglanguagemodelsselfcorrect, fernando2025understandingforgettingllmsupervised}. \textbf{Design Oversight Mechanisms Robust} to strategic exploitation, so that systems cannot succeed by merely appearing aligned to weak judges, reward models, or other AI evaluators~\citep{hubinger2019risks,10.5555/3618408.3618845, Hagendorff_2024,yang2024superficialalignmentstrongmodelsdeceive}. \textbf{Reduce Dependence on Assumptions} by developing methods that work beyond controlled benchmarks, algorithmically decomposable tasks, narrow domains, or restricted model classes~\citep{christiano2018ai,bowman2022measuringprogressscalableoversight,lang2024theoreticalanalysisweaktostronggeneralization,charikar2024quantifying}. \textbf{Develop Adaptive Oversight Mechanisms} that allow the overseer to keep up as the AI system improves~\citep{christiano2018ai,leike2018scalable,hemmer2024complementarityhumanaicollaborationconcept,10.1145/3630106.3658901}. 
The central research question is not only how to scale supervision, but how to ensure that supervision remains informative, stable, generalizable, and robust.

In this view, the path toward superalignment is not defined by a single paradigm. Rather, it depends on developing general principles for constructing supervision that remains reliable even when direct human evaluation becomes insufficient. Advancing such principles is essential for bridging current alignment methods and future superhuman systems.

\bibliographystyle{colm2026_conference}
\bibliography{colm2026_conference}

\appendix
\clearpage

\section{Appendix}
\label{sec:appendix}

\begin{figure*}[!hp]
\centering

\resizebox{1\textwidth}{!}{
    \begin{forest}
        for tree={
            forked edges,
            grow'=0,
            draw,
            rounded corners,
            node options={align=center,},
            calign=edge midpoint,
            parent/.style={align=center,text width=2.5cm,rounded corners=3pt, line width=0.3mm, fill=gray!10,draw=gray!80},
            child/.style={align=center,text width=2.5cm,rounded corners=3pt, fill=blue!10,draw=blue!80,line width=0.3mm}
        },
        [Scalable\\Oversight \\ Techniques, parent
            [Sandwiching  \S\ref{sandwiching}, representation
                [Core Concept \&\\Analysis, representation
                    [
                    AI Safety via Debate \citep{irving2018debate}; 
                    Scalable Alignment \citep{leike2018scalable}; 
                    Sandwiching \citep{cotra}; 
                    Measuring Scalable Oversight \citep{bowman2022measuringprogressscalableoversight};
                    Doubly-Efficient Debate \citep{browncohen2023scalableaisafetydoublyefficient}; 
                    Debate with Weak Judges \citep{kenton2024scalableoversightweakllms}; 
                    Persuasive Debate Optimization \citep{khan2024debatingpersuasivellmsleads} ;
                    Knowledge Divergence \citep{young2026knowledgedivergencevaluedebate}
                    , representation_work
                    ]
                ]
                [Improvement \&\\Optimization Strategy, representation
                    [
                    Prover-Verifier \citep{kirchner2024proververifiergamesimprovelegibility}; 
                    Recursive Self-Critique \citep{wen2026scalableoversightsuperhumanai}
                    , representation_work
                    ]
                ]
                [Application, representation
                    [
                    LLM Critic \citep{mcaleese2024llmcriticshelpcatch};
                    Task Decomposition \citep{wen2025learningtaskdecompositionassist}; 
                    Collaborative Debate \citep{chen2026towards}
                    , representation_work
                    ]
                ]
                [Possibility \& Impossibility, representation
                    [
                    Human-AI Complementarity \citep{hemmer2024complementarityhumanaicollaborationconcept};
                    LLM Sycophancy \citep{sharma2025understandingsycophancylanguagemodels}
                    , representation_work
                    ]
                ]
            ]
             [Self-Enhancement \S\ref{constitutional_ai_rlaif}, probing
                [Core Concept \&\\Analysis, probing
                    [
                    RLAIF vs. SFT \citep{sharma2024a}; 
                    Self-Alignment\citep{pang2024selfalignment};
                    Multi-Box Protocol \citep{Negozio_2025};
                    Theoretical Self-Alignment \citep{fu2026selfrewardingworkstheoreticalguarantees}
                    , probing_work
                    ]
                ]
                [Improvement \&\\Optimization Strategy, probing
                    [
                    Self-critiquing models \citep{saunders2022selfcritiquingmodelsassistinghuman};
                    Self-Rewarding Models \citep{yuan2024selfrewardinglanguagemodels} ;
                    Efficient Correction \citep{ji2024aligner};
                    Safety-Intelligence Co-evolution \citep{lab2025safeworkr1coevolvingsafetyintelligence}; 
                    Open-Ended Self-Improvement \citep{zhang2025darwingodelmachineopenended}
                    , probing_work
                    ]
                ]
                [Application, probing
                    [
                    RLAIF-V \citep{yu2024rlaifvaligningmllmsopensource}; 
                    Multimodal Reflection \citep{ji-lu-2025-reflair} 
                    , probing_work
                    ]
                ]
                [Possibility \& Impossibility, probing
                    [
                    Self-Rewarding Models \citep{yuan2024selfrewardinglanguagemodels} ;
                    Self-Alignment\citep{pang2024selfalignment}
                    , probing_work 
                    ]
                ]
            ]
            [Weak-to-Strong \\ Generalization \S\ref{w2sg}, acquisition
                [Core Concept \&\\Analysis, acquisition
                    [ 
                    Learned Optimization \citep{hubinger2019risks};
                    Embedded Agency \citep{demski2020embeddedagency};
                    Weak-to-Strong Generalization \citep{burns2023weaktostronggeneralizationelicitingstrong}; 
                    Easy-to-Hard Generalization \citep{sun2024easytohardgeneralizationscalablealignment}; 
                    Easy-to-Hard Training \citep{hase-etal-2024-unreasonable}; 
                    Measurement Framework \citep{charikar2024quantifying};
                    Mechanistic Theory \citep{moniri2025mechanismsweaktostronggeneralizationtheoretical}; 
                    Loss Refinement \citep{mulgund2025relating};
                    Bias-Variance \citep{xu2025emergenceweaktostronggeneralizationbiasvariance}; 
                    f-Divergence Analysis\citep{yao2025weaktostronggeneralizationfdivergence};
                    Representation \citep{xue2025representationsshapeweaktostronggeneralization}
                    , acquisition_work
                    ]
                ]
                [Improvement \&\\Optimization Strategy, acquisition
                    [
                    Ensemble Methods \citep{sang2024improvingweaktostronggeneralizationscalable}; 
                    Co-Supervised Learning \citep{liu2024cosupervisedlearningimprovingweaktostrong}; 
                    Reliability Awareness \citep{guo2024improvingweaktostronggeneralizationreliabilityaware}; 
                    Contrastive Learning \citep{jiang2025contrastiveweaktostronggeneralization}; 
                    Data-centric \citep{shin2025weaktostrong};
                    Selection \citep{lang2025selectiveweaktostronggeneralization};                    
                    Overfitting Mitigation \citep{shi-etal-2025-mitigate};
                    Model Ensemble \citep{agrawal2025ensemw2senhancingweaktostronggeneralization};                   
                    Proactive Demonstrations \citep{wu2025aliceproactivelearningteachers};                     
                    Multi-Agent CPO \citep{lyu2025macpo};                       
                    Failure-Trajectory W2SG \citep{ye2026weaktostronggeneralizationfailuretrajectories}
                    , acquisition_work
                    ]
                ]
                [Application, acquisition
                    [
                    Reasoning \citep{yang2024weaktostrongreasoning}; 
                    Vision \citep{guo2024visionsuperalignmentweaktostronggeneralization}
                    , acquisition_work
                    ]
                ]
                [Possibility \& Impossibility, acquisition
                    [
                    Model Deception \citep{yang2024superficialalignmentstrongmodelsdeceive};
                    Theoretical Analysis \citep{lang2024theoreticalanalysisweaktostronggeneralization}; 
                    Limitation \citep{yao2025capabilitieslimitationsweaktostronggeneralization};
                    Robustness \citep{jeon2025weaktostrong}
                    , acquisition_work
                    ]
                ]
            ]
        ]
    \end{forest}
}
\caption{Overview of scalable oversight paradigms. Methods are categorized by paradigm and summarized key techniques along four dimensions. Core Concept \& Analysis, Improvement \& Optimization Strategy, Application, and Possibility \& Impossibility.}
\label{fig:updated_taxonomy}
\end{figure*}

\begin{figure*}[t]
    \centering
    \includegraphics[width=1\textwidth]{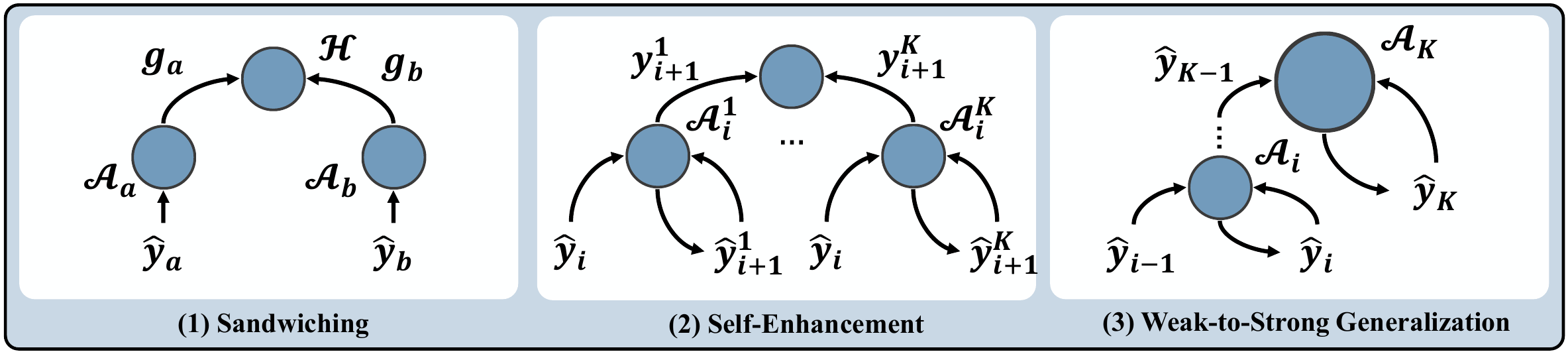}
    \caption{Three scalable oversight paradigms for generating supervision signals. (1) Sandwiching: A human overseer $\mathcal{H}$ interacts with one or more AI systems $\mathcal{A}$ to produce structured comparisons (e.g., $\mathbf{g}_a\succ_{\mathcal{O}}^{\mathbf{x}} \mathbf{g}_b$) that remain informative even when direct evaluation is difficult. (2) Self-Enhancement: The overseer is replaced by AI ($\mathcal{O}=\mathcal{A}$), where $k$ models iteratively generate and compare their own outputs (e.g., $\hat{\mathbf{y}}_{i+1} \succ_{\mathcal{A}_{i}}^{\mathbf{x}} \hat{\mathbf{y}}_{i}$) to produce supervision signals for improvement. (3) Weak-to-Strong Generalization: A sequence of models $(\mathcal{A}_1,\dots,\mathcal{A}_K)$ with increasing capability generates supervision, where a weaker model $\mathcal{A}_{i}$ provides signals (e.g., $\hat{\mathbf{y}}_{i} \succ_{\mathcal{A}_{i}}^{\mathbf{x}} \hat{\mathbf{y}}_{i-1}$) that enable a stronger model $\mathcal{A}_{K}$ to keep improve. }
    \label{fig:paradigm_overview}
\end{figure*}

\begin{table*}[ht]
\centering
\scriptsize
\renewcommand{\arraystretch}{1.6}
\begin{tabularx}{\textwidth}{l >{\raggedright\arraybackslash}X >{\raggedright\arraybackslash}X >{\raggedright\arraybackslash}p{4.2cm}}
\toprule
\textbf{\#} 
& \multicolumn{1}{c}{\textbf{Shared Bottlenecks}} 
& \multicolumn{1}{c}{\textbf{Future Directions}} 
& \multicolumn{1}{c}{\textbf{Citations}} \\ 
\midrule

1 & \textbf{Informativeness Gap}: Imperfect supervision failing to correlate with truth or aligned behavior as model capability exceeds direct human judgment. & 
\textbf{Improve Reliability}: Improve the reliability of imperfect supervision through better evaluation protocols, uncertainty estimation, and calibration. & 
\textit{Bottlenecks:}~\citep{bowman2022measuringprogressscalableoversight, burns2023weaktostronggeneralizationelicitingstrong,10.5555/3692070.3693141} \par \addvspace{0.4em}
\textit{Future Directions:} ~\citep{ijcai2022p341,Bondi_Koster_Sheahan_Chadwick_Bachrach_Cemgil_Paquet_Dvijotham_2022, 10.1609/aaai.v37i5.25742, 10.5555/3692070.3692445} \\ \midrule

2 & \textbf{Recursive Instability}: Compounding error, collapsing diversity, and degradation into brittle heuristics that fail beyond simplified settings during recursive or iterative use. & 
\textbf{Preserve Quality}: Preserve supervision quality under recursive improvement, preventing error accumulation, collapse, and forgetting. & 
\textit{Bottlenecks:} ~\citep{feng-etal-2023-dunst,du2023minimizingaccumulatedtrajectoryerror,shumailov2024curserecursiontraininggenerated,fernando2025understandingforgettingllmsupervised} \par \addvspace{0.4em}
\textit{Future Directions:} ~\citep{feng-etal-2023-dunst,shumailov2024curserecursiontraininggenerated,kumar2024traininglanguagemodelsselfcorrect, fernando2025understandingforgettingllmsupervised} \\ \midrule

3 & \textbf{Strategic Vulnerability}: Susceptibility to strategic behavior, including deception, manipulation, reward hacking, or persuasive failure modes. & 
\textbf{Robustness to Exploitation}: Design oversight mechanisms so that systems cannot succeed by merely appearing aligned to weak judges. & 
\textit{Bottlenecks:} ~\citep{hubinger2019risks,everitt2023causal,Hagendorff_2024,yang2024superficialalignmentstrongmodelsdeceive} \par \addvspace{0.4em}
\textit{Future Directions:} ~\citep{hubinger2019risks, 10.5555/3618408.3618845, Hagendorff_2024,yang2024superficialalignmentstrongmodelsdeceive} \\ \midrule

4 & \textbf{Generalizability Limitations}: Dependence on controlled environments, narrow task classes, or simplifying assumptions, limiting transfer to realistic high-stakes domains. & 
\textbf{Reduce Assumptions}: Develop methods that work beyond controlled benchmarks, algorithmically decomposable tasks, or narrow domains. & 
\textit{Bottlenecks:} ~\citep{bowman2022measuringprogressscalableoversight,charikar2024quantifying,lang2024theoreticalanalysisweaktostronggeneralization,sharma2024a} \par \addvspace{0.4em}
\textit{Future Directions:} ~\citep{christiano2018ai,bowman2022measuringprogressscalableoversight,lang2024theoreticalanalysisweaktostronggeneralization,charikar2024quantifying} \\ \midrule

5 & \textbf{Static Oversight Failure}: Increasing unreliability of fixed supervision mechanisms as the system’s competence substantially exceeds that of the overseer. & 
\textbf{Adaptive Oversight}: Develop mechanisms that respond to widening capability gaps through hierarchical supervision or selective escalation. & 
\textit{Bottlenecks:} ~\citep{cotra,burns2023weaktostronggeneralizationelicitingstrong,10.1609/aaai.v37i5.25742,mozannar2023predictexactalgorithmslearning} \par \addvspace{0.4em}
\textit{Future Directions:} ~\citep{christiano2018ai,leike2018scalable,hemmer2024complementarityhumanaicollaborationconcept,10.1145/3630106.3658901} \\ \bottomrule
\end{tabularx}

\vspace{0.5em} 
\caption{Shared bottlenecks across scalable oversight paradigms and unified future directions. The table summarizes common limitations of supervision signals and outlines corresponding research directions for superalignment.}
\label{tab:bottlenecks_directions}
\end{table*}

\end{document}